\documentclass[10pt,twocolumn,letterpaper]{article}

\usepackage{cvpr}              % To produce the CAMERA-READY version
\usepackage{graphicx}
\usepackage{amsmath}
\usepackage{amssymb}
\usepackage{booktabs}
\usepackage{makecell}
\usepackage{multirow}

\usepackage[pagebackref,breaklinks,colorlinks]{hyperref}

\usepackage[capitalize]{cleveref}
\crefname{section}{Sec.}{Secs.}
\Crefname{section}{Section}{Sections}
\Crefname{table}{Table}{Tables}
\crefname{table}{Tab.}{Tabs.}

\def\cvprPaperID{7156} % *** Enter the CVPR Paper ID here
\def\confName{CVPR}
\def\confYear{2023}

\begin{document}

%%%%%%%%% TITLE - PLEASE UPDATE
\title{Boundary Unlearning}

\author{Min Chen,~~Weizhuo Gao,~~Gaoyang Liu,~~Kai Peng,~~Chen Wang\\
Hubei Key Laboratoryof Smart Internet Technology, School of EIC,\\
Huazhong University of Science and Technology, Wuhan 430074, China\\
{\tt\small \{chenmin7,wzgao,liugaoyang,pkhust,chenwang\}@hust.edu.cn}
% For a paper whose authors are all at the same institution,
% omit the following lines up until the closing ``}''.
% Additional authors and addresses can be added with ``\and'',
% just like the second author.
% To save space, use either the email address or home page, not both
%\and
%Second Author\\
%Institution2\\
%First line of institution2 address\\
%{\tt\small secondauthor@i2.org}
}

\maketitle

\begin{abstract}
The practical needs of the ``right to be forgotten'' and poisoned data removal call for efficient \textit{machine unlearning} techniques, which enable machine learning models to unlearn, or to forget a fraction of training data and its lineage. Recent studies on machine unlearning for deep neural networks (DNNs) attempt to destroy the influence of the forgetting data by scrubbing the model parameters.
However, it is prohibitively expensive due to the large dimension of the parameter space. In this paper, we refocus our attention from the parameter space to the decision space of the DNN model, and propose Boundary Unlearning, a rapid yet effective way to unlearn an entire class from a trained DNN model. The key idea is to shift the decision boundary of the original DNN model to imitate the decision behavior of the model retrained from scratch. We develop two novel boundary shift methods, namely Boundary Shrink and Boundary Expanding, both of which can rapidly achieve the utility and privacy guarantees. We extensively evaluate Boundary Unlearning on CIFAR-10 and Vggface2 datasets, and the results show that Boundary Unlearning can effectively forget the forgetting class on image classification and face recognition tasks, with an expected speed-up of $17\times$ and $19\times$, respectively, compared with retraining from the scratch.
{\renewcommand{\thefootnote}{} \footnotetext[1]
{This work was supported in part by the National Natural Science Foundation of China under Grants 62272183 and 62171189; by the Key R\&D Program of Hubei Province under Grant 2021BAA026; and by the special fund for Wuhan Yellow Crane Talents (Excellent Young Scholar). The corresponding author of this paper is Chen Wang.}}
\end{abstract}

%%%%%%%%% BODY TEXT
\section{Introduction}
\label{sec:intro}

Suppose a company trains a face recognition model with your photos and deploys it as an opened API. Your photos could be stolen or inferenced by attackers via model inversion attack~\cite{fredrikson2015model,khosravy2022model}.
With the increasing awareness of protecting user's privacy, a lot of privacy regulations take effect to provide you the control over your personal data.
For examples,  the General Data Protect Regulation (GDPR) established by the European Union gives individuals ``\textit{the right to be forgotten}" and mandates that companies have to erase personal data once it is requested~\cite{RN245}. 

Beyond the ``right to be forgotten'', data forgetting from machine learning (ML) models is also beneficial when certain training data becomes no longer valid, e.g., some training data is  manipulated by data poisoning attacks~\cite{goldblum2022dataset,peri2020deep}, or outdated over time, or even identified to be mistakes after training.
These practical needs call for efficient \textit{machine unlearning} techniques, which enable ML models to unlearn, or to forget a fraction of training data and its lineage. 

In this paper, we focus on unlearning an entire class from deep neural networks (DNNs), which is useful in realistic scenarios like face recognition: unlearning one's data needs to forget the entire class of one's face images.
As the DNN model retrained from scratch is the optimal unlearned model, early studies try to accelerate the retraining process of deep networks~\cite{bourtoule2021machine, graves2021amnesiac, wu2020deltagrad}, but have to intervene the original training process, which degenerates the model utility and increases the training time. 
A branch of recent researches~\cite{Mehta_2022_CVPR, golatkar2020eternal, golatkar2020forgetting, alex2021ssse} attempt to destroy the influence of the forgetting data by scrubbing the model parameters. For example, the Fisher Information Matrix (FIM) is used to locate the influence of forgetting data at the parameter space~\cite{golatkar2020eternal,golatkar2020forgetting}. 
However, it is prohibitively expensive due to the large dimension of the parameter space.

\begin{figure}[t]
	\centering
	\includegraphics[width=0.72\columnwidth]{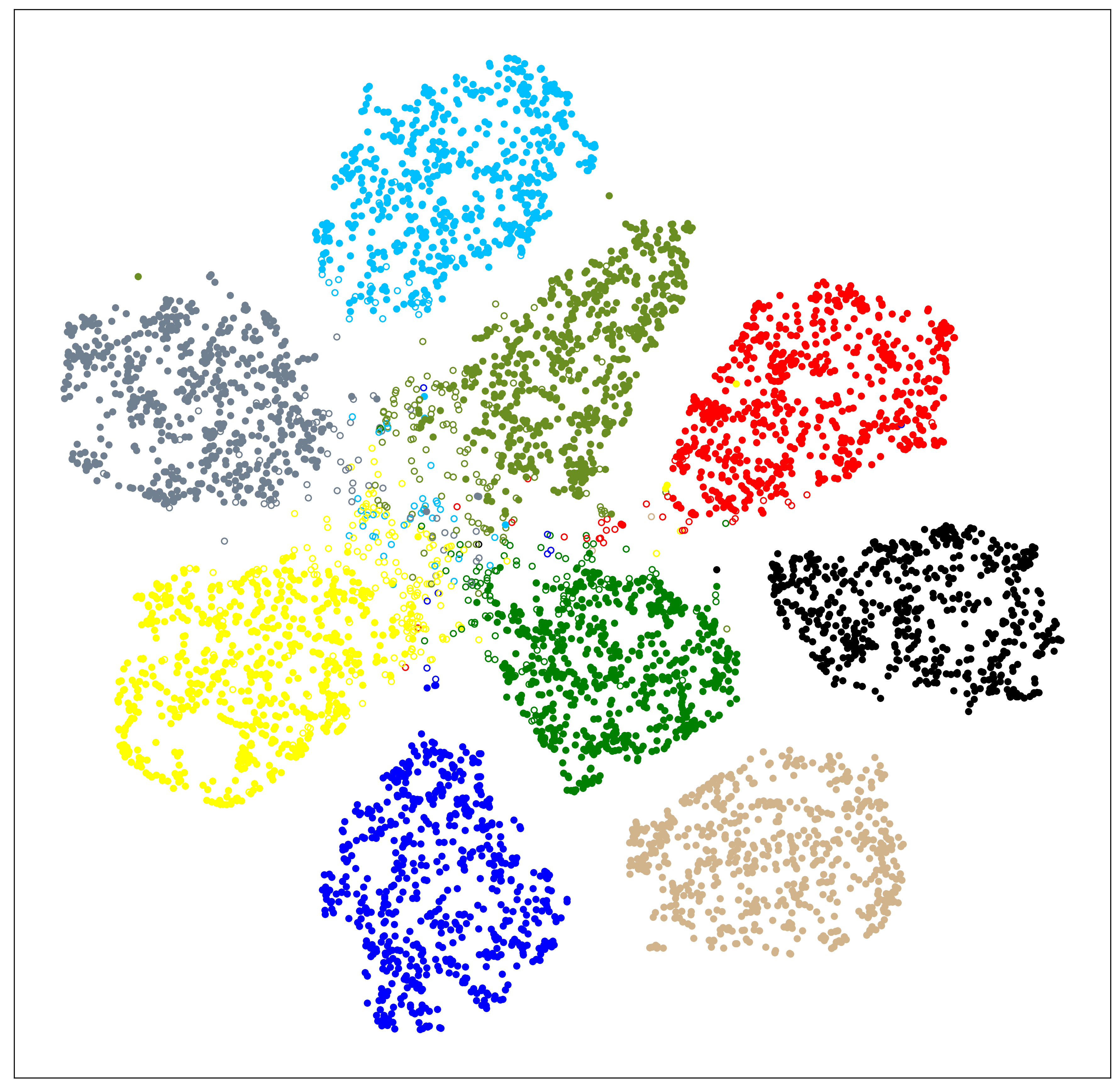} 
	\caption{Key observations from the decision space of the retrained DNN model. The solid dots in different colors represent the remaining samples belonging to different classes and the hollow circles in different colors stand for the forgetting samples predicted as corresponding classes. It can be observed that (1) the forgetting samples spread around the feature space of the retrained DNN model, and (2) most of the forgetting samples move to the borders of other clusters.}\vspace{-5pt}
	\label{fig1_key_observ}
\end{figure}

In order to find an efficient unlearning approach to forget an entire class, we visualize the decision space of the retrained DNN model and discover two key observations (c.f. Figure~\ref{fig1_key_observ}). First, the forgetting samples spread around the decision space of the retrained DNN model, indicating that the decision boundary of the forgetting samples has been broken. Second, most of the forgetting samples move to the border of other clusters; this helps us recall the closest-to-boundary criterion~\cite{nguyen2004active} that samples at the border of cluster in the decision space will probably be predicted with huge uncertainty.

These two observations naturally match the two critical goals of machine unlearning: utility and privacy guarantees. Utility guarantee ensures that the unlearned model should generalize badly on the forgetting data while the prediction performance on the remaining data is maintained. Privacy guarantee means that the unlearned model should not leak any information of the forgetting data. 
Based on our key observations, the utility guarantee can be achieved by only destroying the boundary of the forgetting class but maintaining the boundary of the remain classes, while the privacy guarantee can be accomplished by pushing the forgetting data to the border of other clusters. 

In light of the above ideas, we refocus our attention from the parameter space to the decision space of the DNN model\footnote{Previous unlearning approaches try to destroy the information of the forgetting data by locating the influential parameters directly, while we find that unlearning can be accomplished by manipulating the parameters with the guidance of the decision behaviors of the retrained model.}, and propose \textit{Boundary Unlearning}, a rapid yet effective way to unlearn the forgetting class from a trained DNN model. Boundary Unlearning tries to shift the decision boundary of the original DNN model to imitate the decision behavior of the retrained model. 
To achieve the critical goals, we further introduce two novel boundary shift methods: \emph{Boundary Shrink} and \emph{Boundary Expanding}. The former breaks the decision boundary of the forgetting class by splitting the forgetting feature space into other classes, while the latter disperses the activation about the forgetting class by remapping and pruning an extra shadow class assigned to the forgetting data.        

We summarize our major contributions as follows:
\begin{itemize}
	\item We propose Boundary Unlearning, the first work to unlearn an entire class from a trained DNN model by shifting the decision boundary. Compared with existing studies, Boundary Unlearning neither costs too much computational resource nor intervenes the original training pipeline. 
	\item We propose two novel methods, namely, Boundary Shrink and Boundary Expanding, to shift the decision boundary of the forgetting class. Both methods can rapidly achieve the utility and privacy guarantees with only a few epochs of boundary adjusting.  
	\item We conduct extensive experiments to evaluate Boundary Unlearning on image classification and face recognition tasks. The results show that Boundary Unlearning can rapidly and effectively forget the forgetting class, and outperforms four state-of-the-art techniques.
	The code has been released for reproducibility\footnote{\url{https://www.dropbox.com/s/bwu543qsdy4s32i/Boundary-Unlearning-Code.zip?dl=0}}. 
\end{itemize}
%-------------------------------------------------------------------------

\section{Related Work}
Machine unlearning is first introduced on convex models to find some comprehensible indications, but is found more useful and challenging for DNNs.   
Existing unlearning methods for DNNs can be broadly categorized into two groups: retrain acceleration and updating parameters.

\textbf{Unlearning for ML Models.}
Machine unlearning is first proposed in statistical query learning, which makes the trained model forget specific data by transforming learning algorithm into a summation form~\cite{cao2015towards}. 
Obviously, not every ML algorithm can be converted to summation form. So a few works~\cite{guo2020certified, ginart2019making, brophy2021machine} explore the unlearning task on convex models and try to find some constructive indications with solid theoretical bases. 
However, they cannot generalize to DNNs due to the non-convex nature of the loss functions. 

\textbf{Retrain Acceleration for DNN Models.}
The DNN model retrained from scratch with the remaining data is a naive yet optimal unlearning method, but it is expensive in terms of the cost of training time and resources. 
A few approaches are thus proposed to accelerate the retrain process. An approach called SISA adopts the idea of ``divide and conquer" and proposes to split and mark the training data to train some sub-models and then aggregates them into the final well-trained model~\cite{bourtoule2021machine}.  Graves \etal~\cite{graves2021amnesiac} achieve retrain acceleration by saving the gradient information of each training batch during the original training phase and unlearn the forgetting data by subtracting the gradient update of the specific batch. Similarly, Wu \etal~\cite{wu2020deltagrad} propose Deltagrad, which saves and subtracts the gradient of each forgetting sample by leveraging Quasi-Newton method. More recently, rapidly retrain has also been applied in federated learning to tackle the unlearning problem~\cite{wang2022federated, liu2021federaser, liu2022right}.
Since all these approaches have to intervene the original training pipeline, they may inevitably hurt the utility of the DNN model and cannot fit the needs of practical Machine Learning as a Service (MLaaS) platforms~\cite{weng2022mlaas,hanzlik2021mlcapsule}.

\textbf{Updating Parameters for DNN Models.}
Another branch of researches attempt to unlearn a DNN model by updating its parameters according to the forgetting data. Golatkar \etal~\cite{golatkar2020eternal} propose Fisher Forgetting to scrub the weights clean of information about the forgetting data, by applying noisy Newton update on parameters of the DNN model. To locate the influence of the forgetting data at the parameter space, Fisher Forgetting adopts the Fisher Information Matrix (FIM), restricted to the remaining data, to compute the specific noise to destroy the information of the forgetting data.  However, it is prohibitively expensive to compute FIM due to the large dimension of the parameter space. 
%Specifically, it is hard for us to locate which part of parameters are most influenced by the forgetting data. 
To alleviate this issue, Peste~\etal~\cite{alex2021ssse} approximate FIM in an empirical way, which requires a single gradient computation per sample. 
Considering that the null-space of weights with similar activations may disable the added noise when it destroys the information of the forgetting data, 
Golatkar \etal~\cite{golatkar2020forgetting} propose NTK (Neural Tangent Kernels) forgetting~\cite{jacot2018neural}  by transforming the trained DNN model into a linearized NTK model and adding the same noise to a linearized version of the trained model.
Nevertheless, the computational complexity of these methods is not that scalable as the increasing of data size. Also, we find the specific noise of these methods may still hurt the utility of the remaining data. One potential reason would be that the influence of the forgetting data at the parameter space cannot be estimated exactly due to the poor interpretability of DNNs.

\begin{figure*}[t]
	\centering
	\includegraphics[width=1.9\columnwidth]{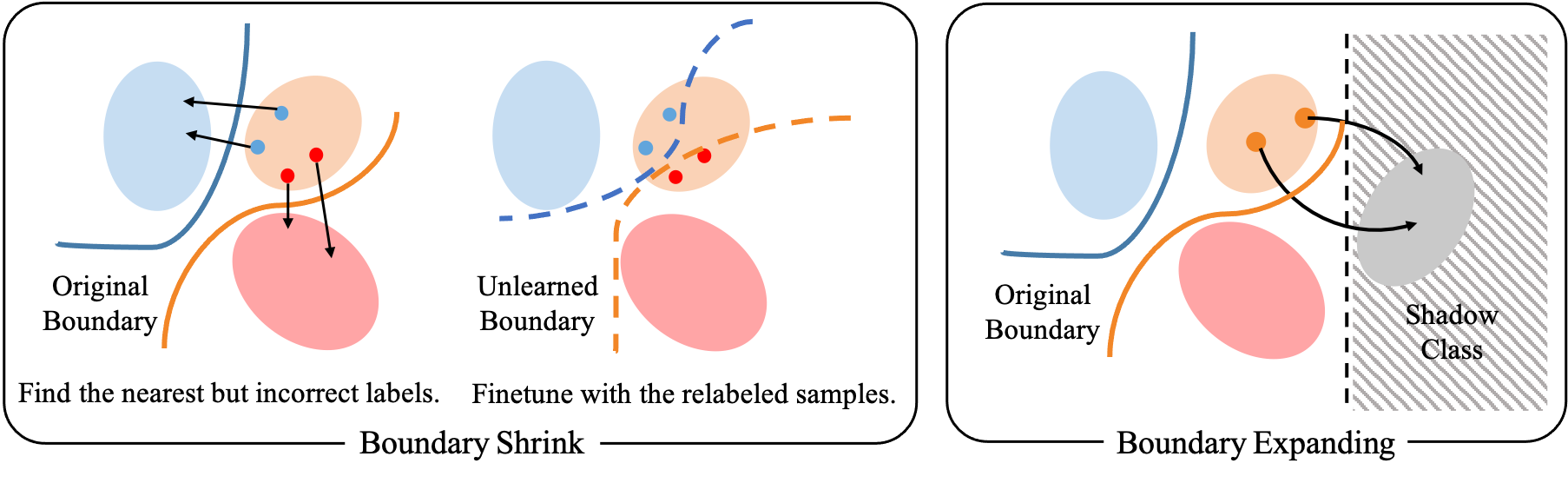} 
	\caption{Illustrating the key ideas of Boundary Shrink and Boundary Expanding. The ovals in different colors represent features of samples in different categories and the dots represent the training samples of the forgetting class. Both Boundary Shrink and Boundary Expanding can destroy the decision boundary of the forgetting class.}
	\label{fig2}
\end{figure*}

\section{Preliminaries and Notation}
We define the notion of machine unlearning in the context of supervised classification with DNNs.
Let $\mathcal{D}=\{\mathbf{x}_i, \mathbf{y}_i\}_{i=1}^{N}\subseteq \mathcal{X}\times\mathcal{Y}$ be the training dataset 
where $\mathbf{x}_i\in\mathcal{X}$ denotes one input and $\mathbf{y}_i\in\mathcal{Y}$ denotes the corresponding class label.
$\mathcal{Y}=\{1, . . ., K\}$ denotes the label space where $K$ is the total number of classes. We denote $\mathcal{D}_f \subseteq \mathcal{D}$ as a subset of training data as the forgetting data, and  $\mathcal{D}_r=\mathcal{D}\setminus \mathcal{D}_f$ as the remaining training data containing the information we expect to retain.
In this work, we primarily focus on the case where $\mathcal{D}_f$ consists of the samples of an entire class.

Let the original DNN model trained on $\mathcal{D}$ be represented by $f_{\mathbf{w}_0}$ parameterized by $\mathbf{w}_0$. Given an input $\mathbf{x}$, $f_{\mathbf{w}_0}(\mathbf{x})$ is the logit output by the trained model on the input $\mathbf{x}$. Given  $f_{\mathbf{w}_0}$, we aim to unlearn the information of  $\mathcal{D}_f \subseteq \mathcal{D}$ from  $f_{\mathbf{w}_0}$ by updating the parameters $\mathbf{w}_0\rightarrow \mathbf{w}'$, where $\mathbf{w}'$ represents the updated parameters obtained by unlearning methods. 
Note that in the problem of unlearning, we expect that the unlearned model $f_{\mathbf{w}'}$ is as similar to the retrained model $f_{\mathbf{w}^*}$ as possible, and $f_{\mathbf{w}^*}$ is retrained on the remaining data $\mathcal{D}_r$ as the optimal unlearning model. 

In Boundary Unlearning, we mainly focus on decision boundary of each class pair $(i,j)$, which can be defined as: $\mathcal{B}^{(i,j)}\triangleq\{\mathbf{x}|f^i(\mathbf{x})=f^j(\mathbf{x})=\max\limits_{k}f^k(\mathbf{x})\}$, where $i, j \in \{1,, . . ., K\}$ denote labels of the class pair and $f^i(\mathbf{x})$ denotes the $i$-th element of the output of DNN on input $\mathbf{x}$. Let $\mathbf{x}_f\in\mathcal{D}_f$ denote a forgetting sample and the label of the forgetting class is $t$. Then, the prediction of $\mathbf{x}_f$ on the retrained model $f_{\mathbf{w}^*}$ will behave as $\mathop{\arg\max}\limits_{k}f^k_{\mathbf{w}^*}(\mathbf{x}_f)\neq t$. 
%In Boundary Unlearning, we want imitate the decision boundary of retrained model $\mathcal{B}^{(i,j)}_{\mathbf{w}^*}$.

\section{Proposed Methods}
Observing that imitating the decision behavior of the retrained model $f_{\mathbf{w}^*}$ is able to accomplish the utility and privacy guarantees of machine unlearning, we transfer our attention from parameter space to decision boundary and design two strategies to shift decision boundary in Boundary Unlearning as follows.

\textit{Boundary Shrink} splits the decision space of the forgetting class by the constraint:  $\mathop{\arg\max}\limits_{k}f^k_{\mathbf{w}'}(\mathbf{x}_f)=y_{nbi}$.

\textit{Boundary Expanding} disperses the activation about the forgetting data by the constraint: $f^t_{\mathbf{w}'}(\mathbf{x}_f)\approx0$.

\subsection{Boundary Shrink}

An intuitive boundary shifting method is to finetune the trained DNN model $f_{\mathbf{w}_0}$ with randomly labeled forgetting data, but this will also shift the boundary of the remaining class randomly, leading to the degeneration of utility the on remaining data. 
As can be observed from Figure~\ref{fig1_key_observ}, most of the forgetting samples are predicted as specific classes, instead of random classes. It reminds us to discover the similarity between forgetting samples and samples of other classes in the feature space.  

Motivated by recent advances of adversarial attacks~\cite{rosenberg2021adversarial,xu2020adversarial}, which can generate adversarial examples across the nearest decision boundary, we thus propose a neighbor searching method to identify the nearest but incorrect class labels to guide the way of boundary shifting (c.f Figure~\ref{fig2}).
Our neighbor searching shows us the direction of shifting the decision boundary of forgetting samples, and can attain the nearest but incorrect labels by predicting \textit{the cross samples} on original  model.
Then we can assign the labels of the cross samples to their corresponding forgetting samples.
In this way, finetuning the trained model $f_{\mathbf{w}_0}$ with all reassigned samples will shrink the boundary of the forgetting class in a precise direction. 

More formally, we assume the original model $f_{\mathbf{w}_0}$ is optimized by the loss function $\mathcal{L}$, where $\mathcal{L}$ can be any standard loss functions, such as cross-entropy. $\mathbf{w}_0$ is the optimal solution of the original model:
\begin{equation}
	{\mathbf{w}_0 = \mathop{\arg\min}\limits_{\mathbf{w}}\sum_{(\mathbf{x}_i,\mathbf{y}_i)\in\mathcal{D}}\mathcal{L}(\mathbf{x}_i,\mathbf{y}_i,\mathbf{w}) }
\end{equation}

Then, we find the nearest but incorrect label for each forgetting sample with our neighbor searching method, by adding noise vector whose elements are equal to the sign of the elements of the gradient of the loss function.
Note that this is somewhat similar to adversarial attacks, but the difference is that we do not need to seek the imperceptible noise (actually, we set a relatively large noise bound), so our  method will be much faster.
% (see Appendix for more experimental results).

Given an initial forgetting sample $\mathbf{x}_f$ and a noise bound $\epsilon$, our neighbor searching updates its cross example using:
\begin{equation}
	{\mathbf{x'}_f = 
		\mathbf{x}_{f} + \epsilon\cdot\mathop{\rm sign}(\bigtriangledown_{\mathbf{x}_{f}} \mathcal{L}(\mathbf{x}_{f}, \mathbf{y}, \mathbf{w}_0)) }
\end{equation}

Once we get the cross samples of all forgetting samples, the nearest but incorrect labels $\mathbf{y}_{nbi}$ can be predicted by the original model $f_{\mathbf{w}_0}$ by:
\begin{equation}
	{\mathbf{y}_{nbi}\leftarrow \mathop{\rm softmax}(f_{\mathbf{w}_{0}}(\mathbf{x'}_f))}
\end{equation}

To shrink the boundary of the forgetting class, we then finetune the original model $f_{\mathbf{w}_0}$ with the reassigned forgetting samples $(\mathbf{x}, \mathbf{y}_{nbi})\in\mathcal{D}_f$, and obtain:
\begin{equation}
	{\mathbf{w}{'} = \mathop{\arg\min}\limits_{\mathbf{w}}\sum_{(\mathbf{x}_i,\mathbf{y}_{nbi})\in\mathcal{D}_f}\mathcal{L}(\mathbf{x}_i,\mathbf{y}_{nbi},\mathbf{w}_0) }
\end{equation}

For the utility guarantee, Boundary Shrink deactivates the power of the DNN model on the forgetting class, but barely hurts the generalization performance on remaining classes. The nearest but incorrect labels help to shrink the boundary of forgetting sample from the sides of other classes, which split the decision space of the forgetting class. Compare to finetuning with random labels, Boundary Shrink achieves the privacy guarantee better as well. Finetuning with random labels will make most of the forgetting samples too conspicuous, which expresses as being predicted  with excessive uncertainty. By contrast, Boundary Shrink pushes the forgetting samples close to the new decision boundary, which makes the unlearned model predict on these forgetting samples with low certainty. The new boundary between remaining classes and forgetting class can be formulated as $\mathcal{B}^{(i,t)}=\{\mathbf{x}_f|\mathop{\arg\max}\limits_{k}f^k_{\mathbf{w}'}(\mathbf{x}_f)=y_{nbi}=i\}$.
Hence, attackers cannot inference these forgetting samples from the unlearned model. 

\subsection{Boundary Expanding}
Although Boundary Shrink achieves both utility and privacy guarantees well, it may cost some time during the neighbor searching. So we propose Boundary Expanding which aims to imitate the decision behavior of the retrained model even more quickly. 

In Boundary Expanding, we do not find the nearest but incorrect way to shift the boundary in decision space. Instead, we artificially assign forgetting samples to an extra shadow class of the original model, which will exploit a new area on decision space (c.f. Figure~\ref{fig2}). 
Recall that we start from our observation that most of forgetting samples move to the border of other clusters on the retrained model. This phenomenon means the forgetting data are predicted as other classes with low certainty. As for a single sample, low certainty represents the output vector is more even. 
Based on this phenomenon, we design a boundary expanding and remapping method to disperse the activation about forgetting data of the unlearned model. 

More specifically, we first introduce an extra shadow class to expand the decision space. This is done by adding a neuron at the last layer of the original DNN model. Then, we finetune the expanded model with the forgetting samples reassigned with the shadow class label: 
\begin{equation}
	{\mathbf{w}{'} = \mathop{\arg\min}\limits_{\mathbf{w}}\sum_{(\mathbf{x}_i,\mathbf{y}_\textit{shadow})\in\mathcal{D}_f}\mathcal{L}(\mathbf{x}_i,\mathbf{y}_\textit{shadow},\mathbf{w}_0) }
\end{equation}

This finetuning approach will remap the forgetting samples to the new area of decision space.  After that, the activation about forgetting data will be collected in the new neuron and the classification neuron of forgetting class will be deactivated. Then, we discard this new area by pruning the extra neuron. 
Thus, the new model unlearned by Boundary Expanding will get back to the same size as the original model, and the pruned model will throw away the information about the forgetting data. 

Intuitively, the expanding and remapping operations only move the forgetting samples in the decision space. As Boundary Expanding never assigns any samples to the remaining classes, the activation of the remaining classes will not change much. Thus, the utility of the DNN model to the remaining data is maintained and the utility guarantee can be achieved by Boundary Expanding. 
Moreover, the classification neuron of the forgetting class in the pruned DNN model will be disabled. Also, the activation about forgetting data will be dispersed to the neurons belong to other classes.  
Therefore, attackers will only get fuzzy logits of the forgetting samples when they attack the pruned model.

\begin{table*}[t]
	\centering
	\caption{Comparison of utility guarantee among baselines and Boundary Unlearning.}
	\label{classification accuary}
	\resizebox{0.9\textwidth}{!}{
		{\fontsize{7}{10}\selectfont
			\setlength{\tabcolsep}{2mm}{
				\begin{tabular}{lllllll|ll}
					\hline
					\textbf{Dataset}                                      & \textbf{Metric}                & \textbf{\makecell[l]{Original\\Model}} & \textbf{\makecell[l]{Retrain}} & \textbf{Finetune} & \textbf{\makecell[l]{Negative\\Gradient}} & \textbf{\makecell[l]{Random\\Labels}} & \textbf{\makecell[l]{Boundary\\Shrink}} & \textbf{\makecell[l]{Boundary\\Expanding}} \\ \hline
					\multicolumn{1}{c}{\multirow{3}{*}{CIFAR-10}} & Acc on $\mathcal{D}_r$ &99.97       &100.00               &100.00          &97.16                   &98.49               & \textbf{99.24}                    &98.03                 \\
					\multicolumn{1}{c}{}                          & Acc on $\mathcal{D}_f$ &99.92       &0.00               &0.22          &7.84                   &10.40               & \textbf{5.94}                    &8.96                 \\
					\multicolumn{1}{c}{}                          & Acc on $\mathcal{D}_{rt}$ &84.83              &85.74               &86.50          &80.42                   &81.81               & \textbf{83.13}                    &81.07          \\    
					\multicolumn{1}{c}{}                          & Acc on $\mathcal{D}_{ft}$ &81.20              &0.00               &0.10          &6.50                   &7.50               & \textbf{5.94}                    &7.00                 \\ \hline
					\multirow{3}{*}{Vggface2}                     & Acc on $\mathcal{D}_r$ &99.94         &100.00         &99.52          &96.57                   & \textbf{98.89}          &98.57               &98.20                 \\
					& Acc on $\mathcal{D}_f$ &98.57         &0.00           &0.00          &2.85                   &4.29           & \textbf{1.54}                &4.22                 \\
					\multicolumn{1}{c}{}                          & Acc on $\mathcal{D}_{rt}$ &98.87         &99.06          &99.96          &99.58                   &95.14          & \textbf{99.72}               &97.12          \\    
					\multicolumn{1}{c}{}                          & Acc on $\mathcal{D}_{ft}$ &97.14         &0.00           &5.52          &7.26                   &2.86           & \textbf{0.87}                &1.41          \\ \hline
				\end{tabular}
	}}}
\end{table*}

\section{Performance Evaluation}

\subsection{Experimental Settings}
\noindent\textbf{Datasets.}
We conduct experiments on CIFAR-10~\cite{krizhevsky2009learning} and Vggface2~\cite{cao2018vggface2} datasets, as in~\cite{golatkar2020eternal, golatkar2020forgetting,tarun2021fast}, to test unlearning performance on image classification task and face recognition task. \\
%\textbf{Models.}
%We use All-CNN~\cite{springenberg2014striving} and ResNet-50~\cite{he2016deep} as the models for CIRAR-10 and Vggface2 datasets, respectively.\\
\textbf{Baselines.} 
We implement the following baseline unlearning methods for comparisons: \\
\textit{Retrain}: we train the model from scratch with the remaining data as the retrained model. Thus, the retrained DNN model is the optimal unlearned model. 
\\
\textit{Finetune}: we finetune the original model on the remaining training data $\mathcal{D}_r$ with large learning rate. \\
\textit{Random Labels}~\cite{hayase2020selective}: finetune the original model on the random relabeled forgetting data $\mathcal{D}_f$. \\
\textit{Negative Gradient}~\cite{golatkar2020eternal}: finetune the original model on the forgetting data $\mathcal{D}_f$ in the direction of gradient ascent. \\
\textit{Fisher Forgetting}~\cite{golatkar2020eternal}: Fisher Forgetting first locates the most relevant parameters in terms of forgetting data and then scrub them by adding noise.\\
\textit{Amnesiac Unlearning}~\cite{graves2021amnesiac}: amnesiac unlearning is a typical method of rapidly retrain. We need save the updates of parameters during several batches on the original training phase. Then, we can unlearn the forgetting data by subtracting the corresponding updates of parameters.\\
\textbf{Implementations.}
Our methods and other baselines are implemented in Python 3.8 and use the PyTorch library~\cite{paszke2019pytorch}.
All experiments are conducted on a workstation with one NVIDIA GeForce RTX 2070 GPU. 
For CIFAR-10 dataset, we train the All-CNN model~\cite{springenberg2014striving} from scratch for 30 epochs using SGD with a fixed learning rate of 0.1, momentum of 0.9 and batch-size of 64. 
For Vggface2 dataset, we randomly select face images of 10 celebrities to conduct experiments. We obtain the original model by finetuning the pretrained ResNet50 model~\cite{cao2018vggface2,he2016deep} with the training images of the 10 celebrities. The original training parameters are similar to those on CIFAR-10.
For the fine-tune process in Boundary Unlearning we use a learning rate of $10^{-5}$ for 10 epochs. 
%Batch-size is set as 64 for all experiments.  
Note that we use the same forgetting class as the forgetting data for all unlearning methods and the rest of classes are the remaining data. \\
\textbf{Metrics.}
We verify the unlearning performance on both utility and privacy guarantees.\\
For utility guarantee, we utilize four accuracy metrics: \textit{accuracy on the remaining training data} $\mathcal{D}_r$, \textit{forgetting training data} $\mathcal{D}_f$, \textit{remaining testing data} $\mathcal{D}_{rt}$, \textit{forgetting testing data} $\mathcal{D}_{ft}$. The unlearned model is expected to get close accuracy with the retrained model.\\
For privacy guarantee, we construct a simple yet general membership inference attack (MIA) based on~\cite{shokri2017membership} using the output of the unlearned model and test the attack success rate (ASR). MIA aims to infer whether a data record was used to train a target ML model or not~\cite{hu2021membership,rezaei2021difficulty,RN362}, and its ASR is widely used as an evaluation metric to verify the privacy guarantee of an unlearning method~\cite{bourtoule2021machine,chen2021machine,NEURIPS2021_9627c45d,Mehta_2022_CVPR}. Ideally, a forgetting procedure should have the same attack success as a retrained model.

\subsection{Utility Guarantee}
For an effective unlearning method, the unlearned model is expected to contain little information about the forgetting data. So we first present comparison results of the accuracy of models by different unlearning methods on CIFAR-10 and Vggface2 datasets in Table~\ref{classification accuary}, which demonstrate the difference of utility caused by the information of the forgetting data. 
From the results we can observe that all the baselines can erase the information of $\mathcal{D}_f$ to some degree. Finetuning the original model on $\mathcal{D}_r$ with lager learning rate and epoch significantly decreases the accuracy on $\mathcal{D}_f$ (0.0\% from the initial 98.57\% on Vggface2). But the more finetune epoch cost more time, and only finetuning on $\mathcal{D}_r$ fails on pursuing the privacy goal, which suggests that finetuning is not a correct unlearning method (as will be shown in the next section). Negative Gradient and Random Lables perform well on accuracy of $\mathcal{D}_r$ and $\mathcal{D}_{rt}$, which means the information of $\mathcal{D}_r$ is maintained, but they fail to erase $\mathcal{D}_f$ completely (still preserve 10.4\% and 7.84\% accuracy of $\mathcal{D}_f$ on CIFAR-10).  
In brief, these baselines unlearn the information about $\mathcal{D}_f$ insufficiently. 

On the contrary, our methods achieve the utility guarantee efficiently. Owing to the well-organized moving direction, Boundary Shrink unlearns the forgetting class and maintains the information of the remaining classes in a more fine-grained way. As shown in Table~\ref{classification accuary}, Boundary Shrink reduces the accuracy on $\mathcal{D}_f$ (only preserves 5.94\% and 1.54\% on CIFAR-10 and Vggface2) but maintains the accuracy on $\mathcal{D}_r$ (degrades 0.73\% and 1.37\% on the two datasets). Compared with baselines, Boundary Shrink harms the information of $\mathcal{D}_r$ to the least extent and erases $\mathcal{D}_f$ most completely. 
Similarly, Boundary Expanding also maintains the accuracy on $\mathcal{D}_r$ and $\mathcal{D}_{rt}$. However, it leaves more residual information about $\mathcal{D}_f$ than Boundary Shrink (8.96\% and 5.94\% on CIFAR-10), but still less than Random Labels (10.40\% on CIFAR-10). As mentioned before, we take Boundary Expanding as a faster alternative method, which takes a trade-off between performance and time consumption. 

\begin{table}[t]
	\centering
	\caption{Comparison of utility guarantee among Fisher Forgetting and Amnesiac Unlearning on CIFAR-10.}
	\label{classification accuary of Fisher}
	\resizebox{\linewidth}{!}{
		{\fontsize{10}{13}\selectfont
			\setlength{\tabcolsep}{1mm}{
				\begin{tabular}{lcccc}%@{}@{}@{}@{}@{}
					\toprule
					\textbf{Metrics} & Acc on $\mathcal{D}_r$ & Acc on $\mathcal{D}_f$ & Acc on $\mathcal{D}_{rt}$ & Acc on $\mathcal{D}_{ft}$ \\
					\midrule
					Amnesiac Unlearning      & 95.79				  &0.00					   &81.50						  &0.00 \\ 
					Fisher Forgetting & 61.62				  &1.80					   &54.20						  &1.60 \\ 
					\bottomrule
				\end{tabular}	
	}}}
\end{table}

We also run Fisher Forgetting on CIFAR-10, which destroys the information about $\mathcal{D}_f$ by adding a specific noise to parameters, to demonstrate the effectiveness of our methods. The results in Table~\ref{classification accuary of Fisher} show that Fisher Forgetting erases $\mathcal{D}_f$ optimally (only preserves 1.8\% accuracy on $\mathcal{D}_f$) but fails to maintain the utility on $\mathcal{D}_r$ (decreases 38.35\% accuracy). However, both Boundary Shrink and Boundary Expanding achieve forgetting and maintain the information of $\mathcal{D}_r$ well. We also show the utility performance of Amnesiac Unlearning in Table~\ref{classification accuary of Fisher}. Amnesiac Unlearning erases information about $\mathcal{D}_f$ ideally, but still hurts the utility of $\mathcal{D}_r$ as it intervenes the original training process. Most importantly, Amnesiac Unlearning costs too much time and memory space to accomplish unlearning, as will be shown in the following section. 

\begin{figure}[t]
	\centering
	\begin{subfigure}{0.47\linewidth}
		\centering
		\includegraphics[width=1.55in]{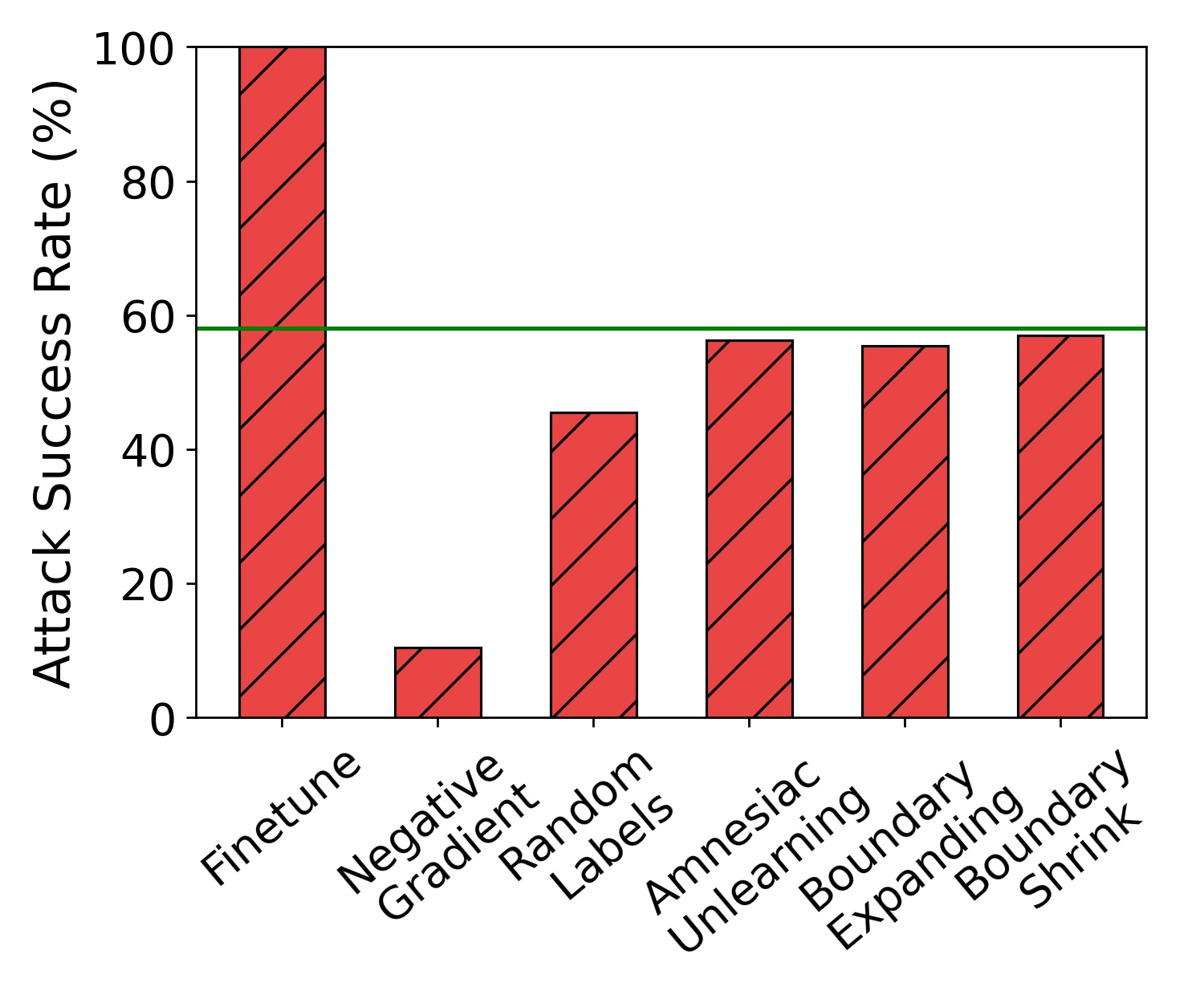} 
		\caption{On CIFAR-10.}
		\label{fig_asr_cifar}
	\end{subfigure}
	\hspace{-2pt}
	\begin{subfigure}{0.47\linewidth}
		\centering
		\includegraphics[width=1.55in]{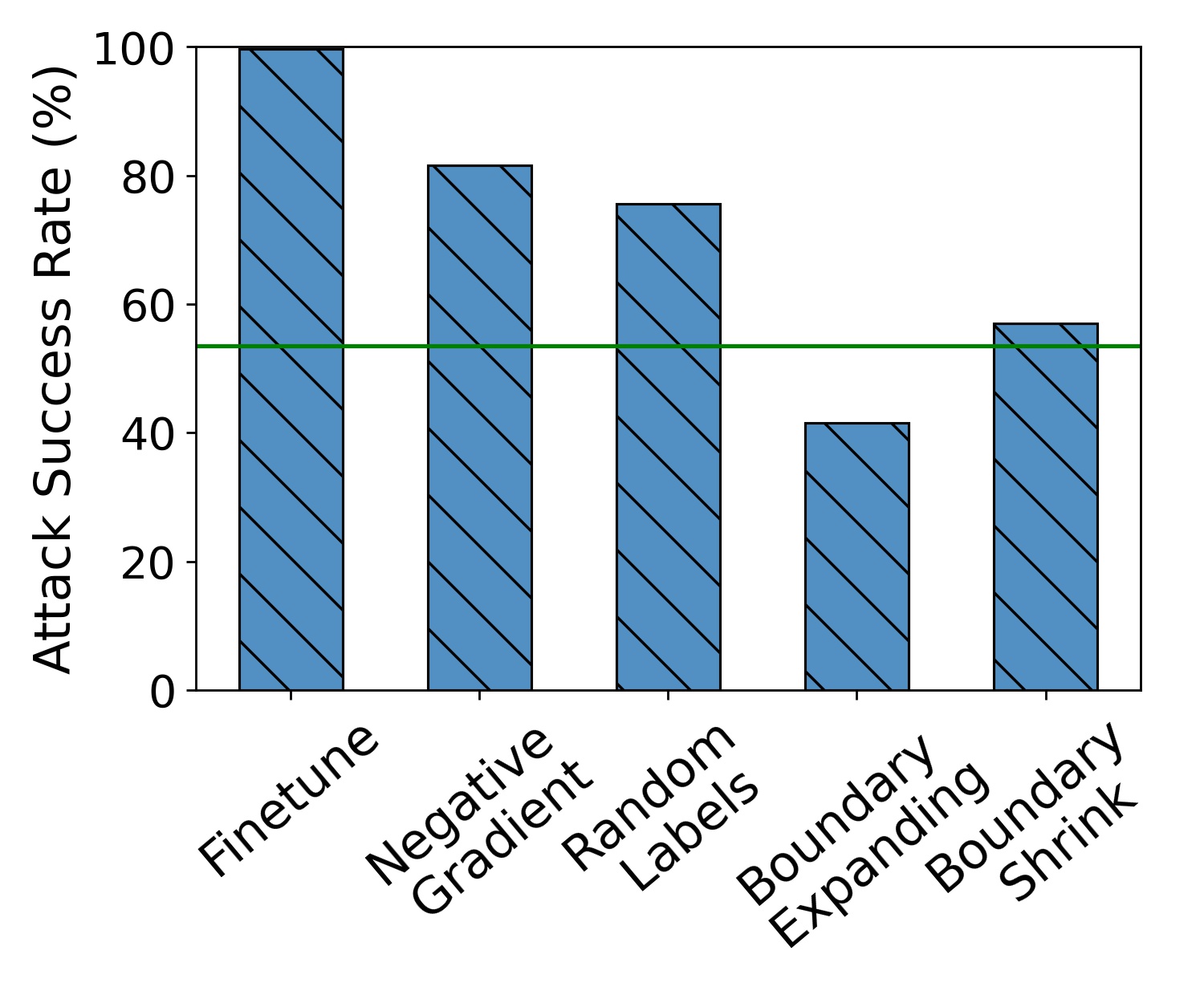} 
		\caption{On Vggface2.}
		\label{fig_asr_face}
	\end{subfigure}
	\caption{The attack success rate of each unlearning method. The green line represents the ASR of the retrained model, and it is expected to be more close to it to reduce the privacy leakage.}\vspace{-5pt}
	\label{fig_asr}
\end{figure}

\subsection{Privacy Guarantee}
Next, we turn to evaluate the privacy guarantee to ensure that the unlearned model by our Boundary Unlearning does not leak any information about the forgetting data. 
We plot the ASR for the unlearning methods on CIFAR-10 and Vggface2 datasets, and the results are shown in Figure~\ref{fig_asr}.  The green line in Figure \ref{fig_asr} represent the ASR of the retrained model and all the unlearning methods attempt to match with it: the closer the better. Here, as an unlearned model gets a closer ASR to 100\%, the less information about $\mathcal{D}_f$ is erased. Contrarily, a quite low ASR (close to 0\%) may represent the Streisand effect which can provide more information about the forgetting data~\cite{golatkar2020eternal}, i.e. all samples of $\mathcal{D}_f$ are predicted as one class.

From the results in Figure~\ref{fig_asr}, we can see that the ASR of Finetune is fairly high on both datasets, which means Finetune barely removes information about forgetting data. For Negative Gradient on CIFAR-10 dataset, it gets a pretty low ASR which is far from the ASR of the retrained model. Random Labels can erase a part of forgetting data but only achieves a relatively low ASR on CIFAR-10. Interestingly, both Negative Gradient and Random Labels have high ASR on Vggface2 dataset, mainly because Vggface2 consists of millions of face images and the basic patterns of faces are more difficult to erase, especially for weaker unlearning methods.  Similar to the test of utility guarantee, we also test Fisher Forgetting, but find that \textit{the ASR of Fisher Forgetting is 0 which represents the severe Streisand effect}. Amnesiac Unlearning obtains an ASR close to the retrained model, as it saves and then subtracts the updates of parameters related to $\mathcal{D}_f$, which makes it more like a rapid retraining model. 

As for our Boundary Shrink/Expanding, they can achieve quite close ASR to that of the retrained model on both datasets. To be more specific, Boundary Shrink achieves better ASR than Boundary Expanding because Boundary Shrink particularly considers the distance between forgetting samples and samples belong to other classes. In a word, Boundary Unlearning methods can achieve a better performance on privacy guarantee.

\begin{figure}[t]
	\centering
	\includegraphics[width=0.67\linewidth]{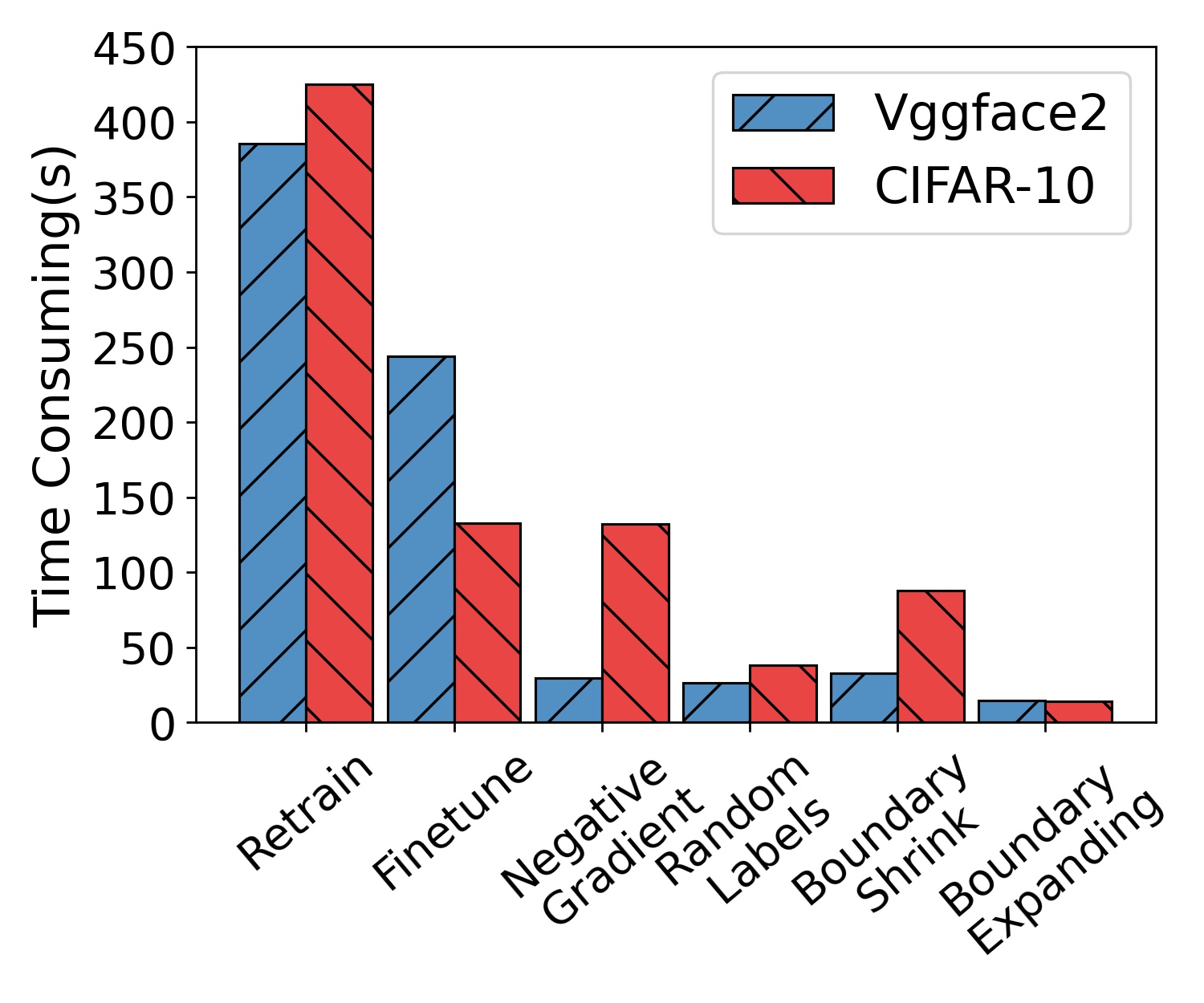}
	\vspace{-2pt}
	\caption{The time consumption of each unlearning method.}\vspace{-5pt}
	\label{fig_time}
\end{figure}

\subsection{Computational Complexity}
In this section, we report the time consumed by each unlearning method to show the computational complexity.  
The major results on CIFAR-10 and Vggface2 are depicted in Figure~\ref{fig_time}. Compared with Retrain, all other unlearning methods of course spent less time, but Finetune takes much more time than other methods,  and it still gets a fairly high ASR in respect of privacy guarantee. As for Boundary Unlearning, both of them can forget the forgetting data in a short time and achieve the privacy guarantee well in the mean time. On the Vggface2 dataset, Boundary Expanding and Boundary Shrink reduce the forgetting time by 26.6$\times$ and 11.7$\times$, respectively. As for CIFAR-10 dataset, Boundary Expanding and Boundary Shrink provide a speed-up of 29.7$\times$ and 4.8$\times$,  respectively. As we discussed before, Boundary Shrink takes a little more time than Boundary Expanding because the generation of cross samples.

Also, we test the time consumption of Fisher Forgetting and Amnesiac Unlearning.\textit{ We do not add the results to Figure \ref{fig_time} because both of them cost extremely long time and the results cannot fit the figure well}. For CIFAR-10 on All-CNN model, Fisher Forgetting costs around $2.7\times10^3$ seconds to unlearn the forgetting class, mainly caused by the huge parameter space and the large amount of training samples. Although Fisher Forgetting can erase the forgetting data perfectly, the time consumption is intolerable. As for Amnesiac Unlearning, the unlearning process consists of parameters subtracting and repairing, which costs around $2.8\times10^2$ seconds. But from our reproduction, we find that Amnesiac Unlearning doubles the time consumption of original training process (costs around $1.0\times10^3$ seconds), which is mainly caused by the operation of saving parameters update. 
Thus, we can see that our Boundary Unlearning can remove the forgetting data effectively and quickly.

\begin{figure}[t]
	\centering
	\includegraphics[width=0.8\linewidth]{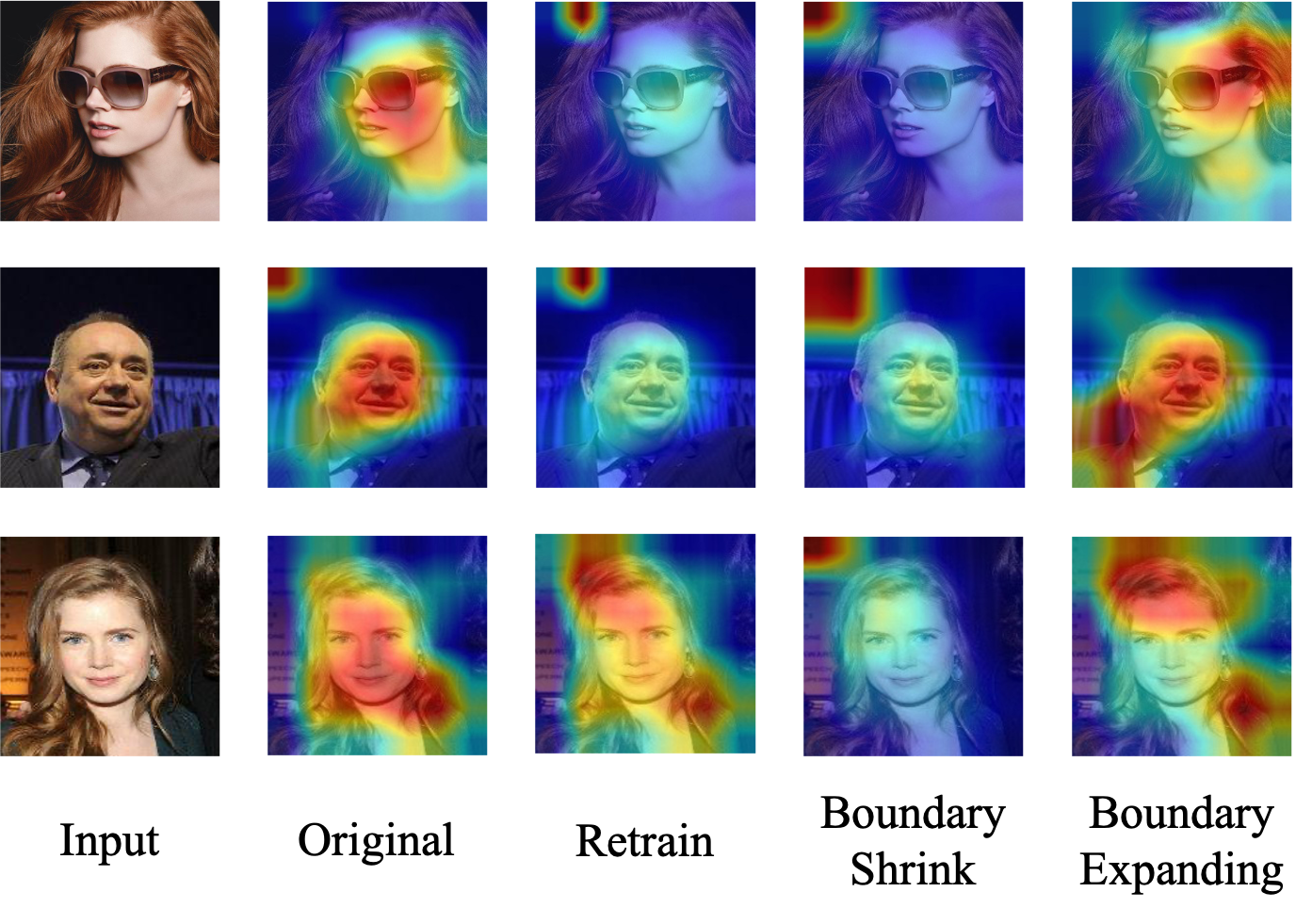}\vspace{-2pt}
	\caption{The attention map of each unlearning method.}\vspace{-5pt}
	\label{fig_map}
\end{figure}

\begin{figure*}[t]
	\centering
	\begin{subfigure}{0.26\linewidth}
		\includegraphics[width=1.55in]{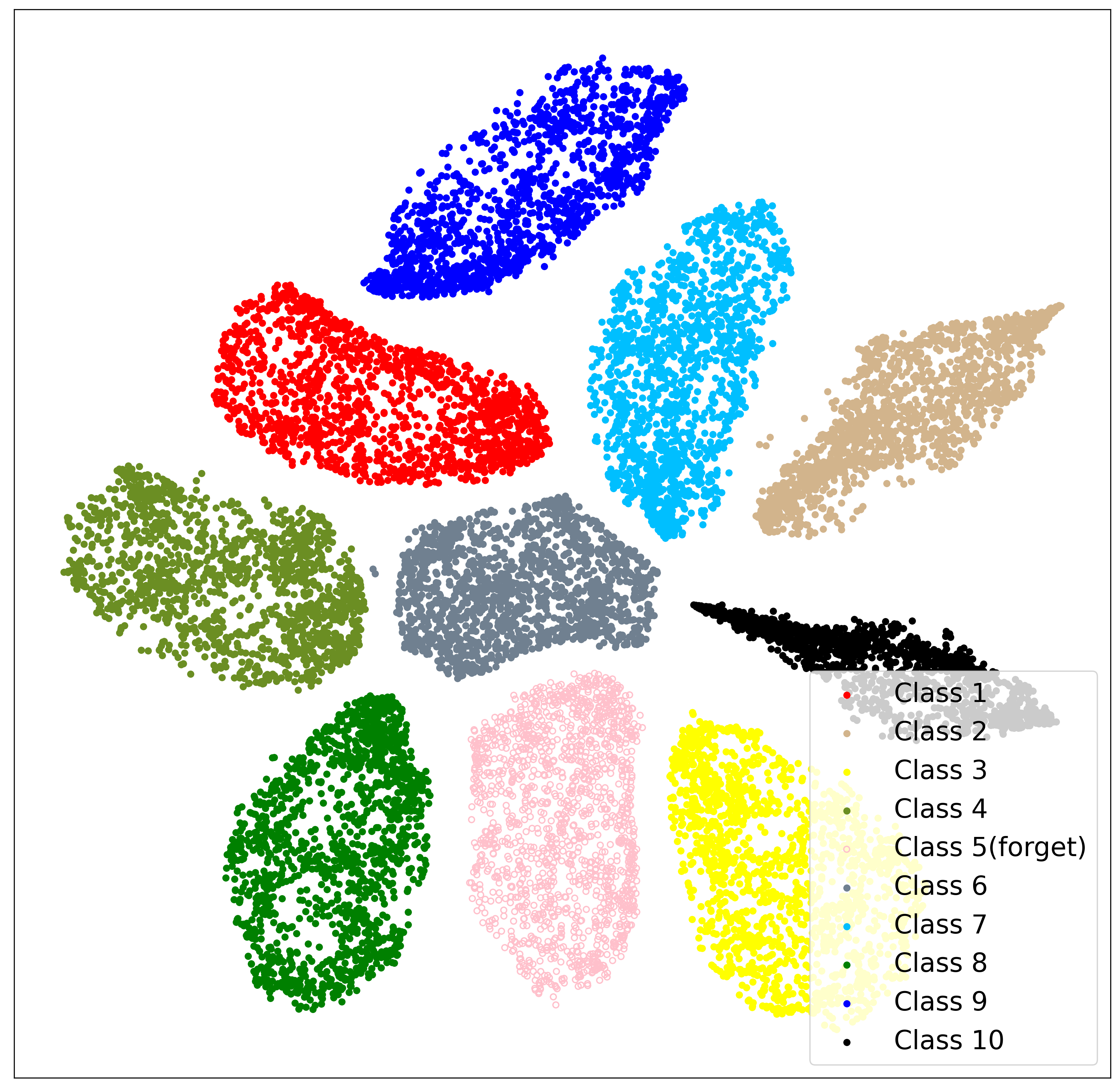}
		\caption{Original}
		\label{Fig_5_(a)}
	\end{subfigure}%\quad
	\begin{subfigure}{0.26\linewidth}
		\includegraphics[width=1.55in]{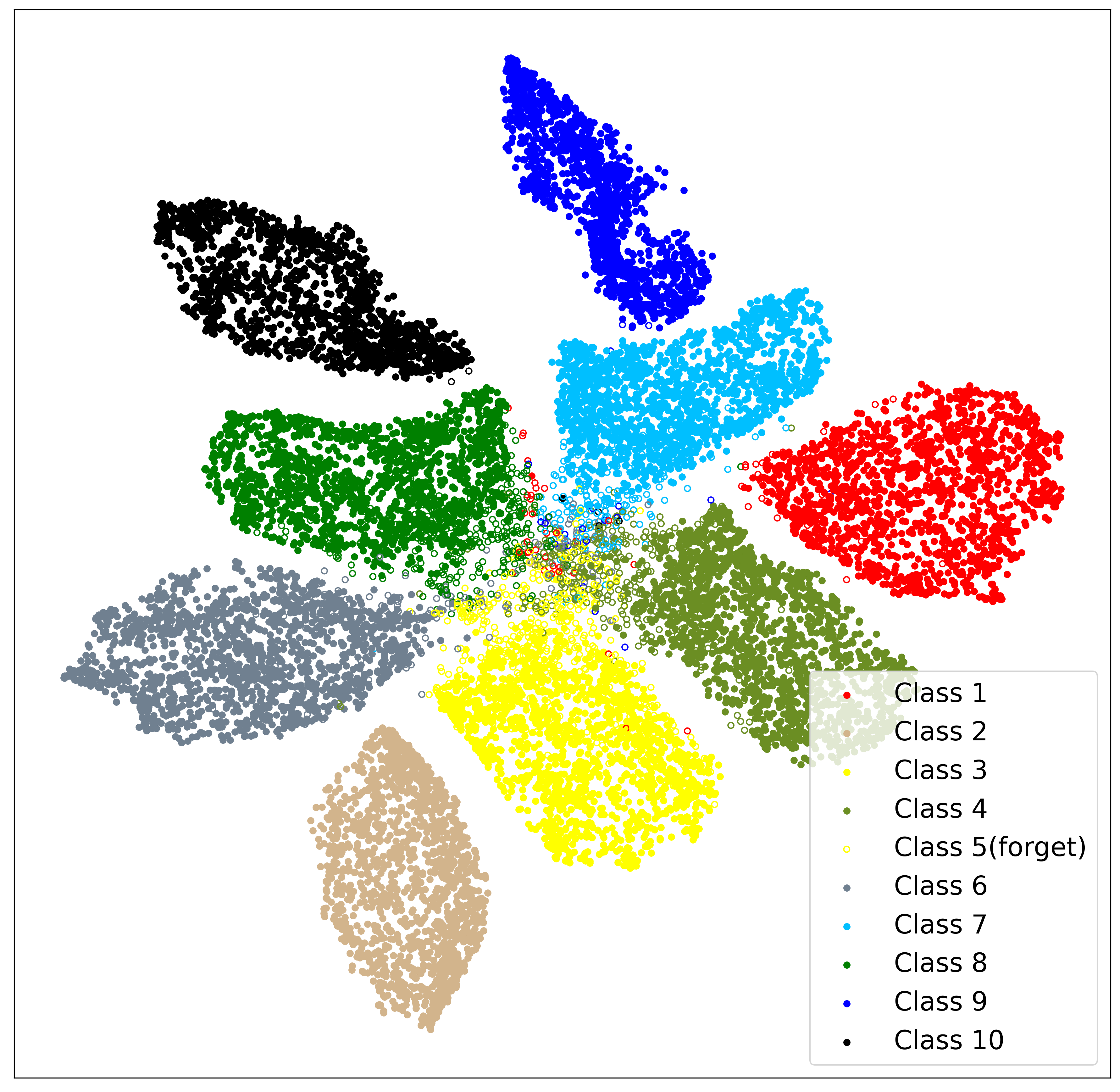}
		\caption{Retrained}
		\label{Fig_5_(b)}
	\end{subfigure}%\quad
	\begin{subfigure}{0.26\linewidth}
		\includegraphics[width=1.55in]{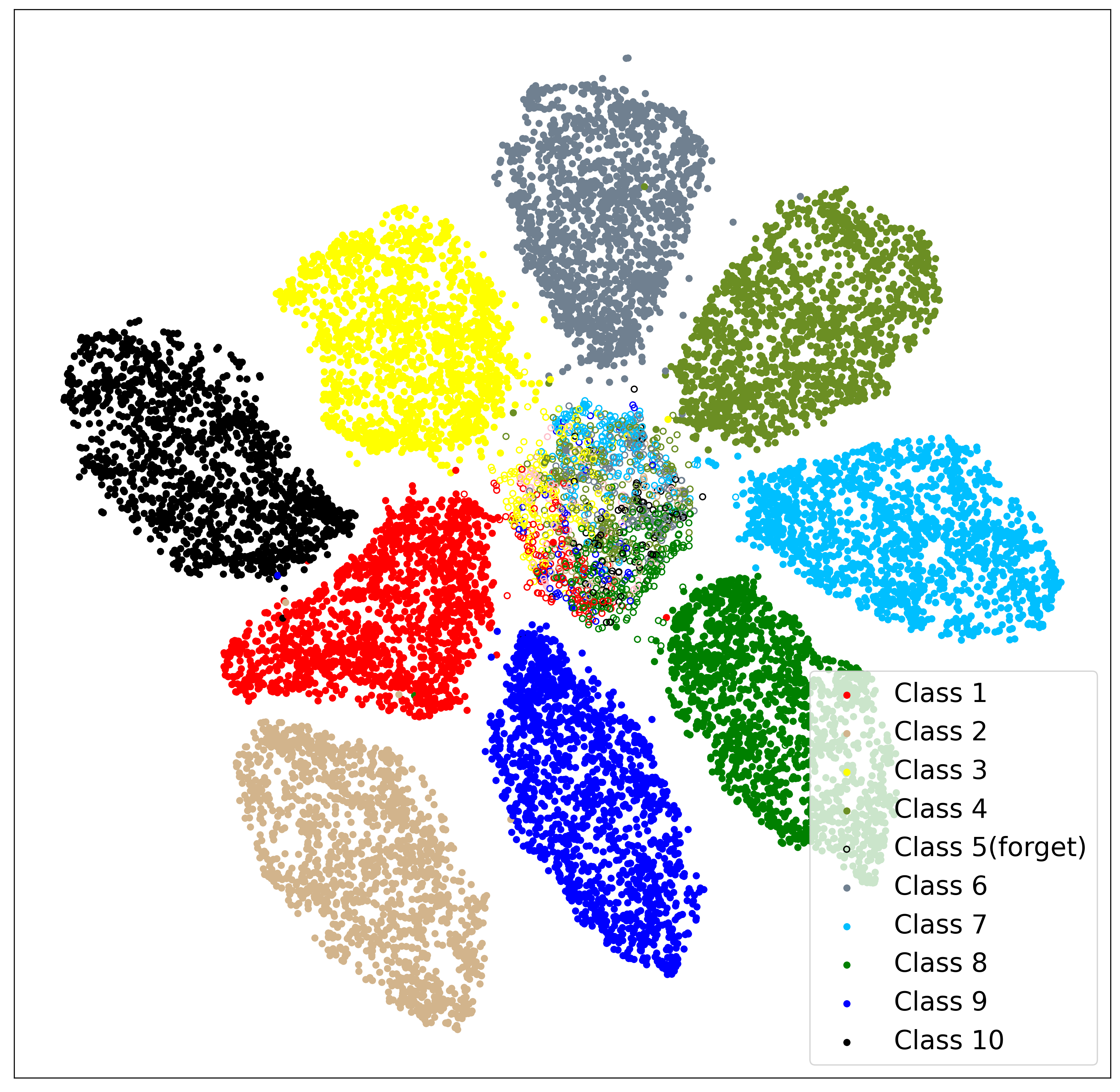}
		\caption{Boundary Shrink}
		\label{Fig_5_(c)}
	\end{subfigure}%\quad
	\begin{subfigure}{0.26\linewidth}
		\includegraphics[width=1.55in]{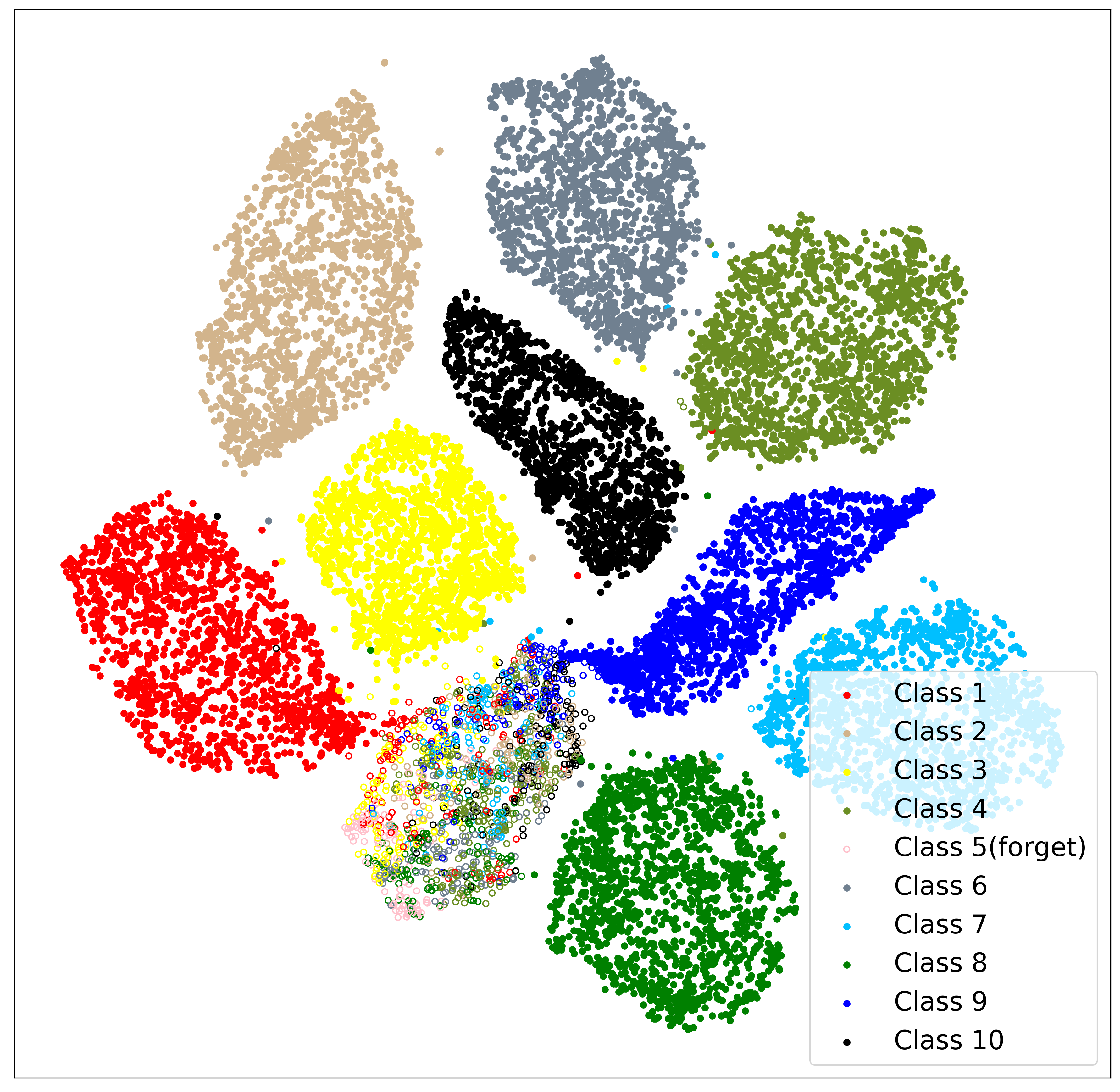}
		\caption{Boundary Expanding}
		\label{Fig_5_(d)}
	\end{subfigure}
	\caption{Visualization of decision space of different methods on CIFAR-10 dataset. The solid dots in different colors represent samples belonging to different remaining classes and the hollow circles stand for the forgetting data. Ideally, we wish the unlearned models act like the retrained model.}
	\label{Fig_5}
\end{figure*}

\begin{figure*}[t]
	\centering
	\includegraphics[width=2.1\columnwidth]{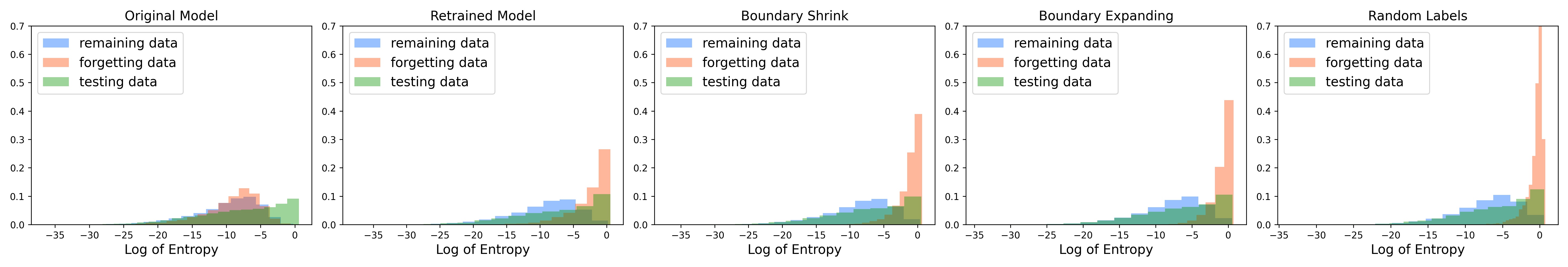} 
	\caption{Distribution of the entropy of model's output on the remaining data $\mathcal{D}_r$,  the forgetting data $\mathcal{D}_f$ and the testing data.}\vspace{-5pt}
	\label{fig6}
\end{figure*}

\subsection{Attention Map}
To make the effect of forgetting more transparent, we plot the attention maps of models on the forgetting data in the Vggface2 dataset before and after applying Boundary Unlearning. The attention map~\cite{selvaraju2017grad} highlights the important areas in the image  for predicting the concept. The columns in Figure~\ref{fig_map} (from left to right) show the focus areas of the original model, the retrained model and unlearned models generated by Boundary Unlearning. We find that the retrained model does not focus on the area of face, and Boundary Unlearning transfers the attention of models from face to background. In particular, the model unlearned by Boundary Shrink only focuses on the background. Although Boundary Expanding fails to transfer the attention completely, it still refocuses on the area outside of face. The results indicate that the output of unlearned model contains hardly any information about the forgetting data. 

\subsection{Visualization of Decision Space}
To make more clear how the decision boundary of the forgetting data shifts, we visualize the decision boundary before and after unlearning in Figure~\ref{Fig_5}. 
We can find that the forgetting data (hollow circles) has been predicted as the nearest classes after applying Boundary Shrink, even the cluster does not entirely spread out like on the retrained model. 
In addition, we find that some hollow circles move to other clusters in Figure~\ref{Fig_5_(c)}. 
These phenomenons reveal that the decision space of the forgetting class is split by its near classes, like on the decision space of the retrained model (c.f. Figure~\ref{Fig_5_(b)}). 
Meanwhile, the clusters of remaining classes still keep compact, which means the model utility to the remaining data is maintained.
In the decision space of the unlearned model generated by Boundary Expanding shown in Figure~\ref{Fig_5_(d)}, the cluster of the forgetting data is pushed away from the center. The forgetting samples are predicted as the remaining classes. Moreover, the clusters of remaining classes are maintained and few remaining data is predicted incorrectly. Therefore, Boundary Unlearning makes the decision boundary of the unlearned model more like that of the retrained model and thus accomplishes the unlearning efficacy. 

\subsection{Distribution of the Entropy of Model Output}
At last, we deep into the distribution of the model's output to figure out why Boundary Unlearning can achieve privacy guarantee. With respect to the distribution of the model's output on the forgetting data, we expect the unlearned models to match closely with the retrained model. On the other side, a great difference between  distributions before and after unlearning will give rise to Streisand effect. 

Figure~\ref{fig6} shows the distribution of entropy of model's output. From the results, we can see that the entropy of outputs on $\mathcal{D}_r$ and $\mathcal{D}_f$ is lower (more confident) than that on the testing data, because both of them are training data of the original model. As the retrained model has not trained on $\mathcal{D}_f$, the entropy of $\mathcal{D}_f$ on it increases evidently (less confident). The distributions of unlearned models generated by Boundary Unlearning are more similar to that of the retrained model. For the distribution on the unlearned model with Random Labels, its  entropy of $\mathcal{D}_f$ is pretty large and vary significantly than that of the retrained model, which may provide even more information about $\mathcal{D}_f$, i.e., all samples of $\mathcal{D}_f$ are predicted as a specific class by the unlearned model, which may make $\mathcal{D}_f$ more prominent to attackers. 

We can also observe that scrubbing $\mathcal{D}_f$ from the original model by Boundary Unlearning makes the distribution of model's output on $\mathcal{D}_f$ more uniform, which means the model is less confident about the prediction. We believe that the change about the distribution is caused by the shifting of the decision boundary.
Boundary Unlearning destroys the boundary of the forgetting class and thus $\mathcal{D}_f$ is predicted as one of its  nearest classes, as shown in Figures~\ref{Fig_5_(c)} and~\ref{Fig_5_(d)}, which means the unlearned model predicts them with low confidence like predicting the testing samples. This will make the attackers harder to inference the membership information about $\mathcal{D}_f$. 

\section{Conclusion}
In this paper, we have presented Boundary Unlearning, the first machine unlearning methodology to remove information of an entire class from a trained DNN by shifting the decision boundary. 
It neither costs too much computational resource nor intervenes the original training pipeline. Extensive experimental results demonstrate its rapid and efficient forgetting performance in both utility and privacy guarantees of unlearning. 
We envision our work as a practical step in machine unlearning towards revealing the relationship between decision boundary and forgetting.

%\newpage
{\small
\bibliographystyle{ieee_fullname}
\bibliography{egbib}
}

\end{document}